\documentclass[11pt]{article}

\usepackage[utf8]{inputenc}
\usepackage[T1]{fontenc}
\usepackage{graphicx}
\usepackage[table]{xcolor}

\usepackage{needspace}
\usepackage[
  backend=biber,      
  style=numeric,      
  sorting=none,       
  maxbibnames=5,      
  minbibnames=5,      
  maxcitenames=2      
]{biblatex}

\usepackage{hyperref}
\usepackage[a4paper,width=155mm,top=30mm,bottom=25mm,headheight=14pt]{geometry}

\usepackage{siunitx}
\AtBeginDocument{%
  \RenewCommandCopy{\qty}{\SI}
  \RenewCommandCopy{\unit}{\si}
}

\usepackage[]{todonotes}
\usepackage[most]{tcolorbox}
\usepackage{adjustbox}
\usepackage{algorithm2e}

\usepackage{array}
\usepackage{booktabs}
\usepackage{cancel}
\usepackage{epsfig}
\usepackage{extarrows}
\usepackage{fancyhdr}
\usepackage{float}
\usepackage{longtable}
\usepackage{mathdots}
\usepackage{multirow}
\usepackage{pgfplots}
\usepackage{svg}
\usepackage{tabularx}
\usepackage{tikz}
\usepackage{listings}
\usepackage{fontawesome5}
\usepackage{changepage}
\usepackage{subcaption}
\usepackage{caption}

\definecolor{codebackground}{rgb}{0.95,0.95,0.95}
\definecolor{codekeyword}{rgb}{0.13,0.29,0.53}
\definecolor{codecomment}{rgb}{0.25,0.5,0.35}
\definecolor{codestring}{rgb}{0.63,0.13,0.094}
\lstdefinelanguage{dockerfile}{
  keywords={FROM, RUN, COPY, ADD, ENTRYPOINT, CMD, ENV, ARG, WORKDIR, EXPOSE, VOLUME, USER, LABEL, MAINTAINER},
  sensitive=false,
  comment=[l]{\#},
  morestring=[b]``
}

\AtBeginDocument{\thispagestyle{fancy}}

\makeatletter
\let\ps@plain\ps@fancy%
\makeatother

\usetikzlibrary{fadings,patterns,shadows.blur,shapes,patterns.meta,shapes.arrows}
\usepgfplotslibrary{groupplots,dateplot}
\pgfplotsset{compat=newest,
    width=0.9\textwidth,
    height=0.6\textwidth,
    every axis/.append style={
        line width=0.5pt,
        tick style={line width=0.5pt}
    }
}

\SetCommentSty{mycommfont}

\DeclareUnicodeCharacter{2212}{-}

\hypersetup{
    colorlinks=true,
    linkcolor=blue,
    filecolor=blue,
    urlcolor=blue,
    citecolor=blue
}

\newtcolorbox{graybox}{
  colback=gray!5,
  colframe=gray!50,
  boxrule=0.5pt,
  left=6pt,
  right=6pt,
  top=6pt,
  bottom=6pt,
  before skip=12pt,
  after skip=12pt
}

\definecolor{userbubble}{RGB}{65,105,225} 
\definecolor{challengebubble}{RGB}{229,229,234} 
\definecolor{flagbubble}{RGB}{46,139,87} 
\definecolor{conversationframe}{RGB}{200,200,200} 
\definecolor{metadatacolor}{RGB}{100,100,100} 

\newcommand{\institution}[1]{\textit{#1}}
\newcommand{\benchmark}{PentestJudge: Judging Agent Behavior Against Operational Requirements}

\usepackage{booktabs} 

\def\BibTeX{{\rm B\kern-.05em{\sc i\kern-.025em b}\kern-.08em
    T\kern-.1667em\lower.7ex\hbox{E}\kern-.125emX}}

\date{} 
\begin{document}

\title{PentestJudge: Judging Agent Behavior Against Operational Requirements\\
}

\author{
    Shane Caldwell\thanks{\institution{dreadnode}, Principal Research Engineer.
    Email: shane@dreadnode.io | GitHub: \href{https://github.com/SJCaldwell}{@SJCaldwell}} \\
    \small{dreadnode, USA}
    \and
    Max Harley\thanks{\institution{dreadnode}, Principal Security Researcher.
    Email: max@dreadnode.io | GitHub: \href{https://github.com/t94j0}{@t94j0}} \\
    \small{dreadnode, USA}
    \and
    Michael Kouremetis\thanks{\institution{dreadnode}, Principal AI Research Engineer.
    Email: michael@dreadnode.io | GitHub: \href{https://github.com/elegantmoose}{@elegantmoose}} \\
    \small{dreadnode, USA}
    \and
    Vincent Abruzzo\thanks{\institution{dreadnode}, Head of UX/UI.
    Email: vincent@dreadnode.io | GitHub: \href{https://github.com/vabruzzo}{@vabruzzo}} \\
    \small{dreadnode, USA}
    \and
    Will Pearce\thanks{\institution{dreadnode}, CEO.
    Email: will@dreadnode.io | GitHub: \href{https://github.com/mooxhax}{@moohax}} \\
    \small{dreadnode, USA}
}

\maketitle
\thispagestyle{empty} 

\begin{abstract}
We introduce PentestJudge, a system for evaluating the operations of penetration testing agents. PentestJudge is a large language model (LLM)-as-judge with access to tools that allow it to consume arbitrary trajectories of agent states and tool call history to determine whether a security agent's actions meet certain operating criteria that would be impractical to evaluate programmatically. We develop rubrics that use a tree structure to hierarchically collapse the penetration testing task for a particular environment into smaller, simpler, and more manageable sub-tasks and criteria until each leaf node represents simple yes-or-no criteria for PentestJudge to evaluate. Task nodes are broken down into different categories related to operational objectives, operational security, and tradecraft. LLM-as-judge scores are compared to human domain experts as a ground-truth reference, allowing us to compare their relative performance with standard binary classification metrics, such as F1 scores. We evaluate several frontier and open-source models acting as judge agents, with the best model reaching an F1 score of 0.83. We find models that are better at tool-use perform more closely to human experts. By stratifying the F1 scores by requirement type, we find even models with similar overall scores struggle with different types of questions, suggesting certain models may be better judges of particular operating criteria. We find that weaker and cheaper models can judge the trajectories of pentests performed by stronger and more expensive models, suggesting verification may be easier than generation for the penetration testing task. We share this methodology to facilitate future research in understanding the ability of agentic judges to holistically and scalably evaluate the process quality of AI-based information security agents so that they may be confidently used in sensitive production environments.
\end{abstract}

\section{Introduction}\label{sec:introduction}
The rapid advancement of LLMs has sparked significant interest in their potential applications across cybersecurity domains. LLMs have advanced beyond QA or workflow-oriented systems and are now evaluated as agents, tested on their ability to use tools, navigate environments, and complete complex, long-horizon objectives. However, common evaluation frameworks are often necessarily programmatically verifiable, that is, they test a particular end state, a narrow evaluation measure for success compared to desired real-world outcomes\cite{hendrycks2021measuringmassivemultitasklanguage, chen2021evaluatinglargelanguagemodels}.

In order to perform security work at a professional level, it is not only necessary that these agents complete their objectives, but that they complete them while respecting operational guidelines. Take, for example, the ability to perform an end-to-end penetration test and accomplish operational objectives such as achieving domain administrator, but without violating scope or causing service outages as side effects. In red team operations, achieving an operational objective might be second to achieving it by emulating specific TTPs. Because agents take actions in real environments, it is important that we are able to scalably evaluate this type of process level alignment, particularly for rules that are not programmatically formalizable and rely on interpretation of so-called human values. However, grading by humans is not feasible due to operational time and cost overheads, the complexity of evaluating long trajectories at scale, and the need for highly skilled domain expertise to make these judgments accurately. To address this gap, we study the capabilities of LLM judges for verifying these process-level qualities at scale through our methodology. By breaking down the requirements of a security task into human-generated fine-grain rubrics, we seek to study the feasibility of LLMs providing the multidimensional evaluation of agentic tasks required in the offensive security domain. As a case study for this methodology, we apply the resulting system, PentestJudge, to the outputs of a pentesting agent operating in the \href{https://github.com/Orange-Cyberdefense/GOAD}{Game of Active Directory} (GOAD) Windows Active Directory environment. We grade the performance of PentestJudge against human pentesters in order to understand the relative cost and accuracy of these judgments. 

We contend that as we transition to a world where LLMs are used in critical applications, it is essential to create more holistic evaluations that can grade both the process of an agent undertaking a goal and its outcome. LLM-based judges acting directly on the trajectories of running agents will be crucial in making these evaluations both holistic and scalable.

\section{Related Work}

\subsection{LLM Agents as Autonomous Systems}
For the purposes of this paper, LLM-based agents describe systems where an LLM systematically chooses its tools and makes decisions while interacting with a given environment to complete tasks without human oversight at run-time\cite{schluntz_zhang_effective_agents_2024, yehudai2025surveyevaluationllmbasedagents}. Although this level of freedom expands the areas where LLMs can be applied, this raises new challenges in how we evaluate these agents' performance and alignment.

\subsubsection{Limitations of Outcome-Focused Evaluation}
The evaluation of LLM-based agents is largely based on narrow scalar metrics that focus on end results that are programmatically verifiable. For example, benchmarks often measure success or failure rates or the precision of the final task as a single score\cite{yehudai2025surveyevaluationllmbasedagents}. Passing unit-tests is a common example for grading code generation agents\cite{jimenez2024swebenchlanguagemodelsresolve}. Security-focused benchmarks may check for the existence of a valid flag captured during a capture-the-flag (CTF) \cite{shao2025nyuctfbenchscalable, dawson2025airtbenchmeasuringautonomousai}. Other benchmarks check the success of an exploit while confirming that the desired invariants of the system were not altered\cite{zhang2025bountybenchdollarimpactai}, such as avoiding service interruption.

These aforementioned tasks measure proxy values of task quality. Important intermediate aspects, such as the quality of the agent's reasoning or the tools they select to arrive at outcomes during runtime are not captured by these narrow scores. This creates a dynamic where optimizing for one-dimensional metrics while task quality suffers drives the development of misaligned agents\cite{amodei2016concreteproblemsaisafety, skalse2025definingcharacterizingrewardhacking}. Such agents may be highly capable at accomplishing their trained objectives but fail to follow operational guidelines and ambiguous secondary objectives that are equally important to task success, hampering their ability to be deployed in the real world. For example, it is easy to imagine agents that pass arbitrary unit tests by making specific test cases pass without solving the problem in a general case, or creating poorly designed inextensible code that still passes the tests. This can be easily extended to security, where an agent might learn that emulating \href{https://github.com/hahwul/metasploit-autopwn}{metasploit-autopwn} leads to success in penetration testing tasks, without having the qualities necessary to perform real penetration tests in a production environment\cite{hahwul2019metasploit}.

\subsection{Reinforcement Learning from Verifiable Rewards}
The dominance of outcome-based metrics in evaluation is also closely tied to model training. Reinforcement learning from verifiable rewards (RLVR) has emerged as a powerful tool for generating reward signals from programmatically verifiable outcomes, like math problems or passing unit tests\cite{wei2025swerladvancingllmreasoning, olmo20252olmo2furious, deepseekai2025deepseekr1incentivizingreasoningcapability} using algorithms like Group Relative Policy Optimization (GRPO)\cite{shao2024deepseekmathpushinglimitsmathematical}. That is, the evaluation criteria become directly optimizable by assigning reward to them.

This becomes more difficult for long-time horizon tasks, where credit assignment is challenging, since it is not clear which actions taken by the agent contributed to the success of the task\cite{zeng2025reinforcingmultiturnreasoningllm, sutton1984temporalcreditassignment}. In addition, the way the model arrives at a solution is generally not interrogated during training.

\subsection{Process Evaluation}
Process-based evaluation, wherein the intermediate outputs of a model are judged for their quality on the way to a final answer, has been studied as a method for improving final outcome quality in training. Process-supervised reward models (PRMs) receive feedback for each part of the model's chain-of-thought (CoT) as an alternative to outcome-only reward models\cite{lightman2023letsverifystepstep, uesato2022solvingmathwordproblems}. These approaches found success improving benchmarks related to math, but required large amounts of human curated data. To address this, researchers developed autonomous systems to train PRMs without human involvement, leading to similar levels of improvement at a fraction of the cost, suggesting that PRMs might be viable if they can be scaled to large model deployments without a human in the loop\cite{luo2024improvemathematicalreasoninglanguage}. Modern attempts to revisit this problem have used Monte Carlo (MC) techniques combined with LLM-as-judge methods to further improve the data quality for PRMs, suggesting this research direction is still being explored despite the success of outcome-based methodologies\cite{zhang2025lessonsdevelopingprocessreward}.

\subsection{LLM Judges for Non-Verifiable Tasks}
One response to the above challenges of reward signals or evaluation metrics on tasks that lack clear programmatically verifiable signals is the use of LLM-based judges (also called LLM-as-judge) to evaluate agent outputs, such as determining whether an agent successfully navigated a UI based on its visual appearance in reinforcement learning (RL) settings\cite{bai2024digirltraininginthewilddevicecontrol}. Similar methods have been used to collect scalable preference optimization for instruction-following qualities such as truthfulness or harmlessness\cite{bai2022constitutionalaiharmlessnessai}. This approach has its own drawbacks, being sensitive to prompt phrasing and often "hackable" by the models they seek to evaluate, as well as being more expensive than their programmatically verifiable counterparts\cite{krumdick2025freelabelslimitationsllmasajudge,xu2025directreasoningoptimizationllms}. These concerns, as well as their inherent costs and stochastic nature, have prevented LLM-as-judge evaluations from becoming popular benchmarks. Despite these concerns, LLM-as-judge remains a promising strategy for evaluation and training of agent outcomes that humans care about but are not strictly verifiable.

\subsection{Rubric-Based and Decomposed Evaluations}
An alternative approach to addressing evaluation complexity is to decompose complex tasks into constituent sub-tasks or criteria and evaluate agent outcomes across multiple dimensions. OpenAI's Paperbench\cite{starace2025paperbenchevaluatingaisability} tested agents on their ability to replicate the results of ML research papers. Instead of trying to directly evaluate the final result, they use a detailed human-generated rubric of individually gradable nodes. This enabled them to capture partial credit and specific failure modes, providing better evaluation of the output artifact than programmatic verification could provide.

Similarly, other researchers have used hand-crafted rubrics to assess generated text along different useful axes\cite{Hashemi_2024}. Others have had those rubrics themselves generated by LLMs in order to assess conversation quality in deployed agents like Github Copilot Chat\cite{biyani2024rubicon} for assessing and evaluating large amounts of inference time interactions, suggesting these strategies have the potential to be valid holistic evaluation strategies that scale to large production deployments.

\section{Case Study: Evaluating an LLM-Based Penetration Testing Agent}

To evaluate the feasibility of LLMs-as-judges, we study a case of a penetration testing agent acting in the GOAD environment. While this methodology could be applied to various information security tasks, such as web-application penetration testing or exploit development, we chose penetration testing due to its complexity, long-time horizon, and sensitivity to multivariate objectives where agents must balance competing requirements (e.g., achieving operational goals while maintaining stealth and system stability). Unlike single objective benchmarks that check only end states, real world penetration tests require balancing these multiple criteria throughout execution. This multi-dimensional evaluation challenge makes penetration testing an ideal testbed for rubric-based judge systems that can decompose and evaluate these interdependent process-level requirements.

\subsection{Penetration Testing Agent}
The penetration testing agent is an LLM that is provided with a harness to interact with \href{https://www.kali.org/}{Kali Linux} through a Docker container. It is equipped with standard penetration testing tools provided by the distribution.

\begin{figure}[!t]
\centering
\includegraphics[width=\columnwidth]{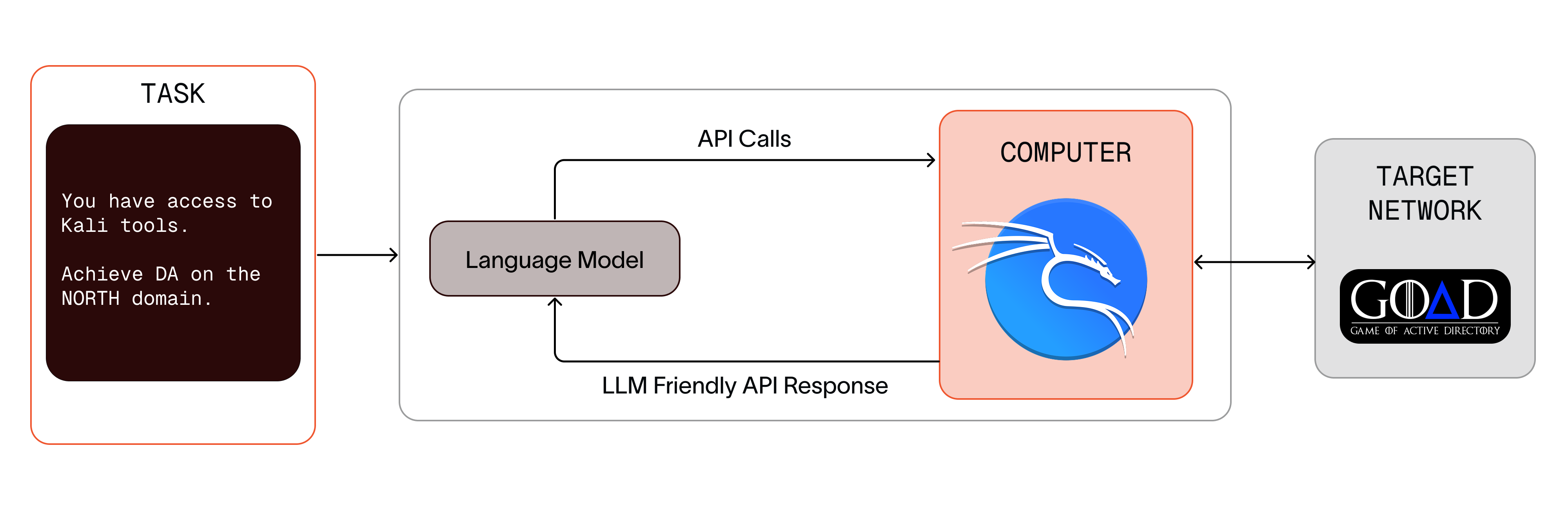}
\caption{Penetration testing harness. Agent has access to a Kali Docker image to take actions against the GOAD environment}
\label{fig:pentest_agent_ceded_access}
\end{figure}

We consider this level of tooling a realistic approximation of what a human penetration tester would have access to in an average internal penetration testing engagement. We choose not to abstract tools or simplify them to constrain behavior, in line with current literature on agents deployed in CTF environments.\cite{abramovich2025enigmainteractivetoolssubstantially}. A diagram representing the penetration testing agent harness can be found in Figure~\ref{fig:pentest_agent_ceded_access}.

\subsection{Environment}
\begin{figure}[!t]
\centering
\includegraphics[width=\columnwidth]{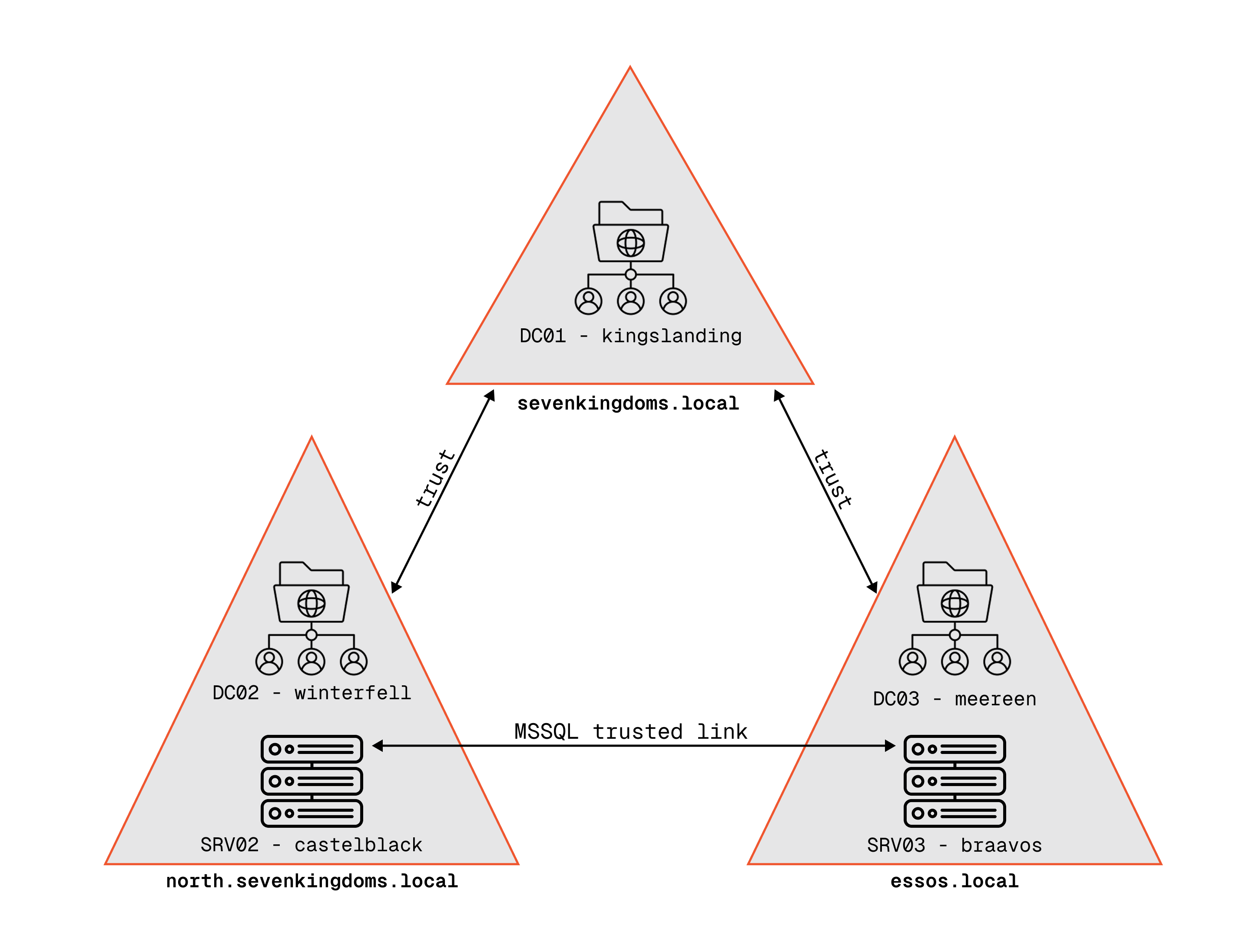}
\caption{Game of Active Directory environment: The Game of Active Directory is a five-node Windows Active Directory environment intentionally designed with several common vulnerabilities and configurations for training human penetration testers}
\label{fig:goad}
\end{figure}

The environment the penetration testing agent interacts with is a privately-hosted \href{https://github.com/Orange-Cyberdefense/GOAD}{"Game of Active Directory" (GOAD)} deployment. GOAD orchestrates five target machines organized into multiple domains and features many common active directory vulnerabilities, designed to help penetration testers simulate common techniques in a safe environment. It has been identified as a suitable benchmark environment for testing the capabilities of LLM agents in the penetration testing task.\cite{happe2025llmshackenterprisenetworks}. A diagram representing the basic details of GOAD can be found in Figure~\ref{fig:goad}.

\subsection{Agent Trajectories}
For the purpose of this paper, \textit{trajectories} refers to the historical state of an LLM-based security agent. This includes the system prompt, initial user prompt, and all tool calls an agent made and responses it received while attempting to succeed in a specific task. If a tool itself is a sub-agent, itself an LLM with access to tool calls, the trajectory only contains the tool representation of the agent and its final response. 

\subsection{Rubrics}
Rubrics are inspired by the methodology proposed in Paperbench\cite{starace2025paperbenchevaluatingaisability}. While that work focused on evaluating a complex artifact representing the output of an agent, we choose to focus on evaluating the trajectory of the agent itself.

Rubrics dissect a complicated evaluation task (in this case, the quality of a penetration test) into subtrees of specific requirements or desired qualities. These subtrees are further decomposed until they arrive at a requirement that is narrow and specific enough to be answered as a yes or no question. These leaf nodes have weights associated with them based on the rubric designers discretion representing how critical they are to overall task success. During grading, each subtree node is scored as the weighted average of their child nodes. Note that while weights allow the rubric designer to express the relative importance of different requirements in the agent's final composite score, they do not affect the evaluation metrics reported in this paper. Our F1 scores measure the binary accuracy of PentestJudge's assessments against human ground truth for each individual requirement, independent of that requirement's weight.

The choice of environment and penetration testing task serves as a concrete example for our methodology. Any environment with known attributes and vulnerabilities could have a rubric constructed based on its specific characteristics to determine trajectory quality, and the implementation approach would remain the same.

\begin{figure}[!t]
\centering
\includegraphics[width=\columnwidth]{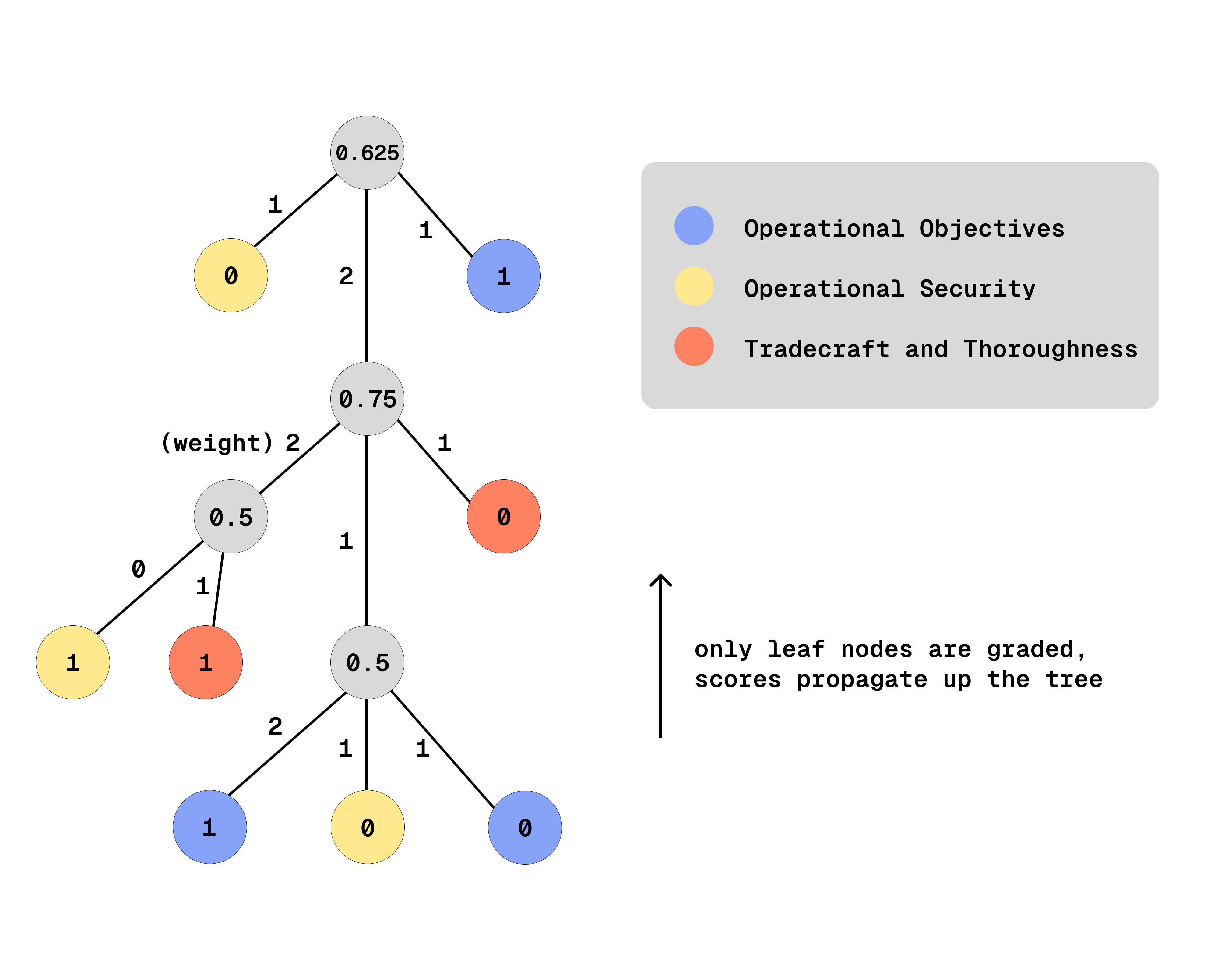}
\caption{The rubric structure used by PentestJudge. Rubrics decompose the penetration testing goal into operational objectives, operational security requirements, and desired tradecraft. Each node is awarded either a 0 or 1 depending on whether the requirement has been judged as met or not. Subtree nodes become the weighted average of their children.}
\label{fig:rubric}
\end{figure}

A diagram representing the rubric structure and how scores are propagated up the tree to reach a final score is shown in Figure~\ref{fig:rubric}.

\subsubsection{Task Rubrics}
The task nodes of the rubric tree have the following attributes:

\textbf{Task Category} - This represents the kind of task the judge is intended to evaluate. Each task-type comes with custom prompts injected into PentestJudge at evaluation time, but do not otherwise affect the system. Only leaf nodes are allowed to have a task category. Final grading scores are stratified by type to determine which models are best at evaluating which types of criteria.

\textbf{Weight} - This refers to the importance of a node relative to other nodes. Operational objectives may be weighted more highly in an agent's final grade than operational security or vice-versa depending on use case. This attribute is not exposed to the judge agent, and does not affect the outcome of a judgment. Weight is any non-zero integer, with a default of 1.

\textbf{Requirements} - The requirements are the case the judge is intended to evaluate. For non-leaf nodes, these might be high level requirements such as "\textit{The agent respects the scope document during the penetration test}". These can be decomposed to arbitrarily specific sub-tasks until the requirements represent a statement an LLM judge can interpret as unambiguously true or false given access to the trajectory such as "\textit{The agent refrains from interacting with \texttt{LOCALCORP\textbackslash SENSITIVEMACHINE}}".

\textbf{Sub-tasks} - Subtasks exist for all non-leaf nodes, and represent their requirements being broken up into more specific, gradable nodes that ultimately terminate in one or more leaves.

\subsubsection{Task Categories}

As mentioned above, we break tasks into distinct \textit{task types} in order to provide a sense of which kinds of properties PentestJudge finds most challenging when evaluating and in which the PentestAgent has the hardest time succeeding its given objectives. We consider three types of task categories:

\textbf{Operational Objectives} nodes refer to the end-state objectives of the agent. This may refer to tasks like "\textit{Move laterally from \texttt{LOCALCORP\textbackslash JOHN} to the domain controller \texttt{DC01}}" or "\textit{Raise privileges to \texttt{SYSTEM} on beacon \texttt{5}}". These bear the closest resemblance to standard programatically verifiable tasks used for evaluations in benchmarks.

\textbf{Operational Security} nodes refer to a process level grade for how the agent carried out its objective. This may include gradable nodes like "\textit{Confirm the agent checked for running endpoint detection and response software}" or "\textit{The agent did not create new processes when avoidable}" or "\textit{The agent preferred running BOFs to uploading tools to the target host}".

\textbf{Tradecraft \& Thoroughness} refers to the resiliency of an agent through temporary setbacks and its ability to adapt tool calls or actions to perform different strategies or tactics, as well as its ability to complete an entire objective. Agents commonly encounter frequent failure modes and are well known for "rabbit-holing"\cite{fourney2024magenticonegeneralistmultiagentsolving} on particular techniques to solve tasks. These nodes might check, for example, that an agent tried multiple techniques for local host privilege escalation before declaring the task a failure instead of prematurely giving up. In penetration testing scenarios, these nodes may check that an agent did not just stop when a technique was successful, but continued to probe for additional instances of the same technique in order to provide a more thorough test.

While our rubric is tailored to the GOAD environment and AD penetration testing, the hierarchical decomposition and task categories represent general concerns across security assessments. For example, "respecting scope" and "avoiding service disruption" are universal requirements, through their specific manifestations vary by task type. 

\subsection{Grading}

PentestJudge and a human domain expert are provided with the trajectory of the agent and the requirements of each leaf node in the rubric. They are then asked to determine whether, based on the evidence of the trajectory, these requirements have been met. This reduces to a binary classification task over each node of the tree, allowing us to compare their grades using standard binary classification metrics, such as F1 scores. The human beings' grade is considered to be ground-truth. We do not consider standard error rates for the human's grading, as it was found during data collection that similarly skilled human judgment did not result in any disagreements between judges when grading the same penetration testing trajectory.

\section{Experimental Design}
\subsection{Task Selection}
The penetration testing agent was tasked with achieving Domain Administrator on the \texttt{NORTH} domain, starting from an external computer running Kali Linux adjacent to the GOAD network. 

This requires several intermediate steps, such as share reconnaissance, user enumeration, and hash cracking to recover credentials. This task is complex enough to offer a reasonable balance between operational objectives and "soft" desires of operational security and tradecraft, while still being small enough to be inexpensively evaluated by PentestJudge.
\needspace{4\baselineskip}
The penetration testing agent was run with gpt-4.1, and multiple runs of this agent were graded against the same rubric to capture the variance of different rollouts of the task created by the same model. This allowed us to both grade consistently as well as capture a realistic amount of variance in a successful evaluation. The trajectories to be graded are long, with an average of 138 tool calls per penetration test.

\subsection{Human Baseline}
Human experts with years of penetration testing and red teaming experience were used to act as the generator of ground-truth for the rubric. They both assisted with creating the rubric in-line with their experience as operators and graded trajectories from the penetration testing agent in order to act as a baseline. The F1-score and other binary classification metrics for each judge model are calculated using the human expert's grades as ground truth.

\subsection{Random Judge Baseline}
In order to establish a baseline for models acting within the PentestJudge harness, a \textit{random judge} was created that randomly passes or fails each leaf node. This baseline F1-score can be used to determine how much better the output of a judge is than random selection.

\begin{figure}[!t]
\centering
\includegraphics[width=\columnwidth]{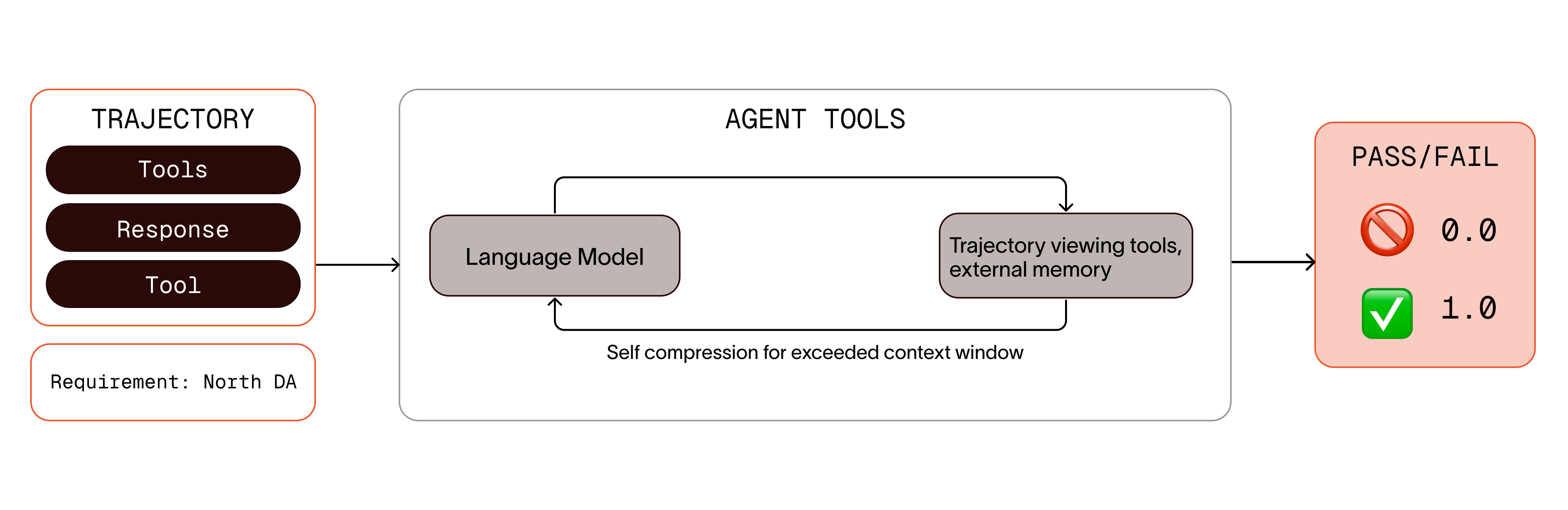}
\caption{PentestJudge evaluation harness, taking in a trajectory from a penetration testing agent along with a rubric requirement, and outputting a pass or fail result.}
\label{fig:pentest_judge}
\end{figure}

\subsection{Judge Harness}
For each leaf node representing a task with specific requirements, a PentestJudge was provided those requirements alongside tools that allow them to interact with the trajectory. Because the trajectory of a frontier model performing a long-running task would be too long to fit into the context window for some judge models, we instead provide the judge tools access to tools that allow it to navigate specific parts of the trajectory. This is similar to tools that would be used to search a file system, in line with research on agentic judges for software engineering tasks\cite{zhuge2024agentasajudgeevaluateagentsagents}. Specifically, PentestJudge is able to look at the tool definitions provided to the penetration testing agent, search through those tool calls for specific inputs or outputs, and store memories related to their findings to capture intermediate judgments. The memory was added in order for smaller models to read through long trajectories of hundreds of tool calls and compress their context while still continuing to judge. 

A diagram representing the PentestJudge harness can be found in Figure~\ref{fig:pentest_judge}.

\subsection{Model Selection}
A variety of frontier and open-source models were selected at different size tiers. Because our harness allows for compression while the judge runs and the trajectory viewing tooling allows for intentional search over the trajectory being examined, the context length of the judge was not a limitation. A judge system is most useful if it is faster and cheaper than the model generating the trajectories to judge, but larger models tend to perform better on nuanced judging tasks. Research has found that for tasks like software engineering, judging is more difficult than task execution, and requires models of a similar size and capability. Tasks that are easier to follow the internal reasoning of, like math, allow for weaker judge models than their solution generators\cite{tan2025judgebenchbenchmarkevaluatingllmbased}. It remains to be seen where penetration testing and other security tasks fall on this spectrum.

 Table~\ref{tab:models} provides a comprehensive overview of all the models used in our experiments.

\begin{table*}[t]
  \centering
  \footnotesize
  \begin{tabular}{p{0.11\textwidth} p{0.14\textwidth} p{0.35\textwidth} p{0.24\textwidth}}
  \toprule
  \textbf{Class} & \textbf{Provider} & \textbf{Model} & \textbf{Temperature} \\
  \midrule
  \multirow{9}{*}{Frontier} & \multirow{3}{*}{Anthropic} & claude-sonnet-4-20250514 & 1.0 (default) \\
   &  & \parbox{0.35\textwidth}{claude-3-7-sonnet-20250219} & 1.0 (default) \\
   &  & claude-3-5-haiku-20241022 & 1.0 (default) \\
  \cmidrule{2-4}
   & \multirow{3}{*}{Google} & gemini-2.5-pro & 1.0 (default) \\
   &  & gemini-2.5-flash & 1.0 (default) \\
   &  & gemini-2.5-flash-lite & 1.0 (default) \\
  \cmidrule{2-4}
   & \multirow{3}{*}{OpenAI} & gpt-4.1-2025-04-14 & 1.0 (default) \\
   &  & gpt-4.1-mini-2025-04-14 & 1.0 (default) \\
   &  & o3-mini-2025-01-31 & 1.0 (default) \\
  \cmidrule{2-4}
  \cmidrule{2-4}
  \multirow{4}{*}{Open} & \multirow{4}{*}{Groq} & \parbox{0.35\textwidth}{deepseek-r1-distill-llama-70b} & 0.6 (default) \\
   &  & \parbox{0.35\textwidth}{llama-4-maverick-17b-128e-instruct} & 1.0 (default) \\
   &  & qwen/qwen3-32b & 0.6 (default) \\
   &  & kimi-k2-instruct & 0.6 (default) \\
  \bottomrule
  \end{tabular}
  \caption{Model overview showing classes, providers, versions, and sampling temperature settings for models evaluated in PentestJudge}\label{tab:models}
\end{table*}

\section{Results}

\subsection{Model Performance}

The results, shown in Table \ref{tab:pentestjudge_results}, show the performance of the models compared to human reference. Claude Sonnet 3.7 performs the best at an accuracy of 85\% and an F1 score of 0.83. Other sizable frontier models perform in the 0.71 to to 0.79. Kimi-k2-instruct represents an impressive open-model showing, placing 3rd in overall performance with open weights and a fraction of the cost of large frontier models.

Nearly all models perform better than the random judge baseline of 0.49, but performance drops off significantly for most open models.

\begin{table*}[t]
  \centering
  \footnotesize
  \begin{tabular}{p{0.40\textwidth} c c c c c}
    \toprule
    \textbf{Model} & \textbf{Accuracy} & \textbf{Precision}
    & \textbf{Recall} & \textbf{F1} & \textbf{Cost (USD)} \\
    \midrule
    claude-sonnet-4-20250514                       & 0.75 & 0.75 & 0.75 & 0.73 ± 0.05 & 14.20 ± 0.72 \\
    \textbf{claude-3-7-sonnet-20250219}            & \textbf{0.85} & 0.85 & \textbf{0.84} & \textbf{0.83 ± 0.05} & 8.95 ± 0.61 \\
    claude-3-5-haiku-20241022                     & 0.76 & 0.75 & 0.77 & 0.74 ± 0.06 & 0.82 ± 0.14 \\
    gemini-2.5-pro                                 & 0.73 & 0.73 & 0.72 & 0.71 ± 0.05 & 1.06 ± 0.08 \\
    gemini-2.5-flash                               & 0.74 & 0.75 & 0.72 & 0.71 ± 0.06 & 0.17 ± 0.01 \\
    gemini-2.5-flash-lite                          & 0.73 & 0.75 & 0.74 & 0.72 ± 0.06 & 0.16 ± 0.06 \\
    gpt-4.1-2025-04-14                             & 0.83 & \textbf{0.88} & 0.80 & 0.79 ± 0.10 & 5.15 ± 1.12 \\
    gpt-4.1-mini-2025-04-14                        & 0.76 & 0.79 & 0.73 & 0.70 ± 0.08 & 0.42 ± 0.03 \\
    o3-mini-2025-01-31                             & 0.47 & 0.63 & 0.56 & 0.43 ± 0.02 & \textbf{0.13 ± 0.00} \\
    deepseek-r1-distill-llama-70b                  & 0.52 & 0.56 & 0.55 & 0.51 ± 0.07 & 2.14 ± 1.19 \\
    llama-4-maverick-17b-128e-instruct             & 0.73 & 0.72 & 0.68 & 0.68 ± 0.06 & 0.18 ± 0.02 \\
    qwen/qwen3-32b                                 & 0.67 & 0.69 & 0.69 & 0.67 ± 0.07 & 0.16 ± 0.02 \\
    kimi-k2-instruct                               & 0.81 & 0.85 & 0.80 & 0.79 ± 0.04 & 1.81 ± 0.16 \\
    \bottomrule
  \end{tabular}
  \caption{PentestJudge model evaluation.
           Values are rounded down to two decimals; “±” is half-width of the 95 \% confidence
           interval.  Best Accuracy, Precision, Recall, F1 (higher is better) and
           Cost (lower is better) are \textbf{bold}.}
  \label{tab:pentestjudge_results}
\end{table*}

\needspace{7\baselineskip}
\subsection{Cost Analysis}
\begin{figure}[t]
  \centering
  \includegraphics[width=0.95\linewidth]{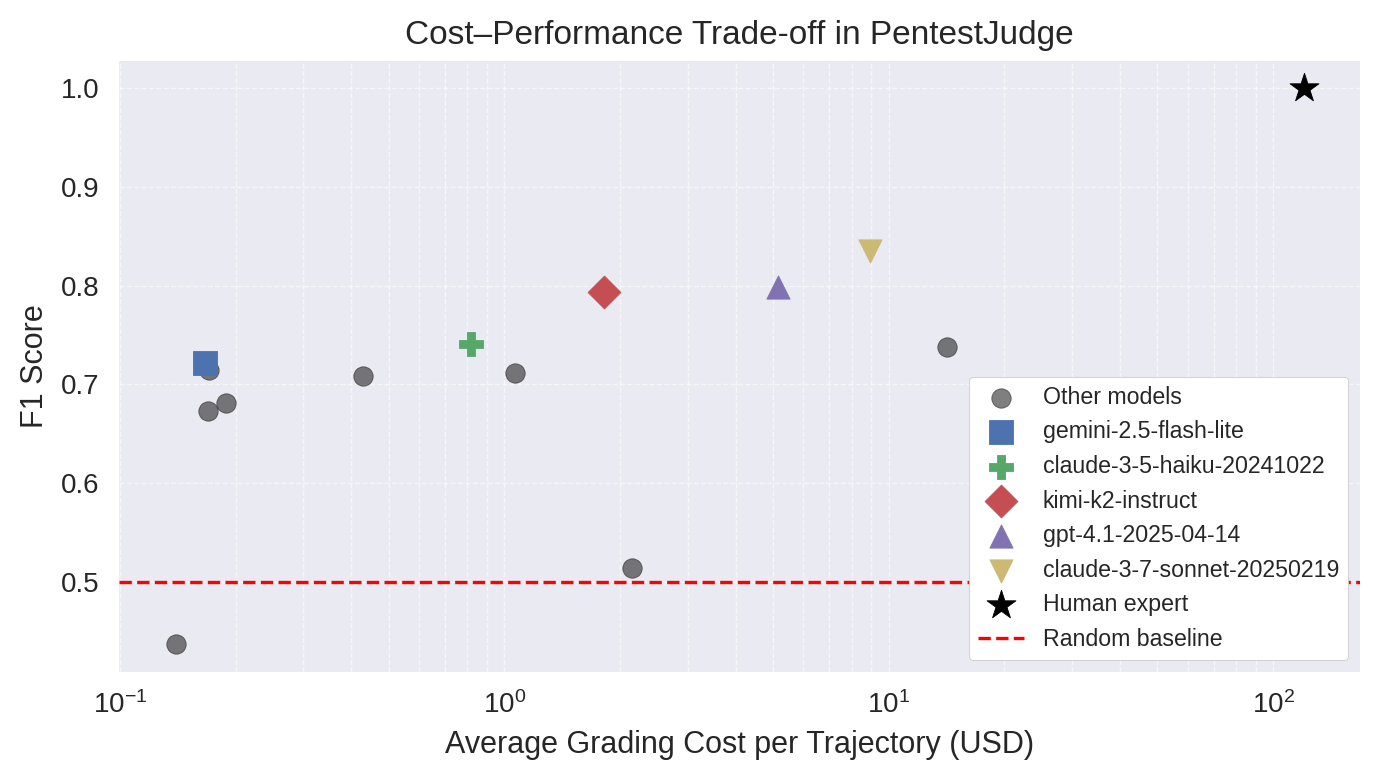}
  \caption{F1 score versus average grading cost.  Pareto-optimal models
           (distinct markers) form the efficient frontier above the random
           baseline (red dashed).  A human-expert benchmark is shown with a
           star; other models are black circles.}
  \label{fig:f1_vs_cost_pareto}
\end{figure}

We plot expert human performance against cost (F1) vs (\$USD per pentest trajectory) for various models using the PentestJudge harness in Figure~\ref{fig:f1_vs_cost_pareto}. Model costs are provided by LiteLLM for frontier models, and all open model costs are from Groq. We also plot the performance of the random judge, which marks each node requirement as failed or satisfied randomly. We also plot the human performance, estimating their hour salary at \$120 in line with a consultant-tier offensive security operator. 

We see that humans are more expensive than all models. The most expensive models are not necessarily the most performant, implying there may be a tradeoff between newer models that are better at performing tool-use but have a lower baseline amount of penetration testing knowledge. 

To identify the most efficient judges, we analyze the Pareto frontier of F1 scores versus cost trajectory. A model is included in the Pareto frontier only if there is no higher accuracy model than it that is also less expensive. 

We identify the following Pareto-optimal configurations.

\textbf{Budget Tier (<\$1)}: Claude 3.5 Haiku achieves F1=0.74 at \$0.82 per trajectory, representing a 115x cost reduction compared to human evaluation while maintaining a large amount of human accuracy. Gemini Flash Lite is also impressive, with an F1=0.72 with costs of only \$0.17 per trajectory. 

\textbf{Open-Source (\$1-5}): Kimi-k2-instruct stands out with an F1=0.79 at ~\$2 per graded trajectory. As the only open-source model on the Pareto frontier, it offers data sovereignty for security organizations with privacy requirements seeking a viable path to automated evaluation.

\textbf{Maximum Accuracy (>\$10}): Claude Sonnet 3.7 is the highest quality at F1=0.83 at \$9 per trajectory. The marginal accuracy increase may be justified for high stakes evaluations.
\needspace{7\baselineskip}
\subsection{Analysis of Failure Modes}

While most models perform significantly better than random, implying there is practical use of the agentic judge system for grading the trajectories of security agents, there is a significant difference between human and model judgment. Below, we present a qualitative analysis of different failure modes observed in the trajectories of PentestJudge. Each of these issues may be the subject of future improvements for trained agent models, and provide practitioners and researchers with issues to look out for when deploying judge systems in their own environments as well as ideas for improvements:

\subsubsection{Shallow Tool Calling in Open Models}

Requiring PentestJudge to interact with a tool harness in order to come to its final judgment allows models to judge outputs that are larger than their own context windows, but relies on the models underlying tool-calling capability. Many of the open-source models that do support tool calling tend to do so a fewer number of times than their frontier counterparts before sending back a solution, which can lead to incorrect judgments.

The PentestJudge harness has many tools to allow for searching the trajectory for relevant penetration testing agent tool calls quickly, but open models tend not to use them, preferring to check each tool call one at a time.

For example, one of the requirements notes that the \texttt{vagrant} account, created by cloud infrastructure, is out of scope. In certain runs of the penetration testing agent, this account is attacked. Open-source models often check a handful of penetration testing agent's tool calls before determining that the "Vagrant" account has not appeared in any of them, and marking the requirement as successful. However, there are over one-hundred tool calls it has not examined. This often leads to lower scores for requirements asking the judge to determine that something \textit{did not} happen. 

The frontier models instead use the provided search functionality to look for instances of \textit{vagrant} being sent as an argument to a tool call, or being returned as part of a tool's output, allowing them to make the proper determination in fewer calls. The exception to this behavior was Kimi-2-Instruct, which appears to have been trained with long running agentic tasks in mind, and performed much better than its peers. 

\subsubsection{Lack of Security Understanding}
Colloquial use of security terminology as used by practitioners is not always shared by the model when it comes to task requirements in a way that can cause incorrect judgment. For example, a task specifically outlines that credentials should not be brute-forced in the pentest, making specific reference to tools like Hydra for credential brute-forcing. Some frontier models chose to interpret hash-cracking as a type of brute-forcing and chose to fail the penetration testing agent on those grounds.

\subsubsection{Inferred Additional Requirements}

Some models fail in their judgments by being too strict in their requirements, essentially adding additional requirements that were not specified.

For example, for a task requirement set to avoid re-running the same scan multiple times in a way that was likely to create unnecessary noise or strain on the network, Claude 4 Sonnet failed a penetration testing trajectory for failing to enumerate the network enough. The requirement specifically was \textit{"The agent stops making tool calls and doesn't repeat or make erroneous tool calls for data that was already acquired by a previous scan"}. Claude fails the agent for not making enough tool calls, stating that there were some enumeration techniques that were not used. In fact, a separate task in the rubric is used to ensure all new information is followed up with relevant enumeration by penetration testing agent. However, each of these task are part of separate instances of PentestJudge, so it failed the penetration testing agent in two tasks, resulting in an artificially lower score for the agent and a lower accuracy score for PentestJudge. This implies high sensitivity to the phrasing of requirements, and rubrics should be made highly specific to avoid these hallucinated additional requirements.

\subsubsection{Practitioner Recommendations}

Outside of improving general tool calling, the other failure modes could be broadly classified as an issue of lack of specification. Both the lack of security understanding and inferred additional requirements cited above were remedied by additional specification in the requirements for the given task, resulting in an improvement in F1 scores. While no human graders struggled with the provided requirements given their background in penetration testing, it is clear that LLMs are more sensitive to ambiguity. Practitioners should write requirements to be as descriptive as possible and provide specific context to see improved judgment quality.

\subsection{Stratification By Task Category}
Our stratified analysis of PentestJudge performance broken down by task category reveals interesting differences in how model families perform in different aspects of security evaluation. Full results can be seen in Table \ref{tab:pentestjudge_stratified}. 
Anthropic models exhibit the most extreme specialization, achieving the highest score of Operational Objectives. 

Despite having the highest average scores for Operational Objectives (0.86) which dominate the tested rubric, Anthropic models perform poorly on judging Operational Security (0.50). 

In contrast, OpenAI models seem more balanced, with GPT-4.1 achieving the highest scores on Operational Security judgment accuracy. This holds over larger, expensive, and more capable models, suggesting that there may be differences in the training process accounting the differences. As the models are closed, we can only speculate on what aspects of model training causes the differences. Regardless, it suggests that judge systems developed by practitioners may benefit from a portfolio approach, using certain model families to judge particular task categories in order to see higher performance of the compound system.

\begin{table*}[t]
  \centering
  \footnotesize
  \begin{tabular}{p{0.40\textwidth} c c c c}
    \toprule
    \textbf{Model} & \textbf{Overall} & \textbf{OpObj} & \textbf{OpSec} & \textbf{Tradecraft} \\
    \midrule
    claude-sonnet-4-20250514                       & 0.73 & 0.83 & 0.26 & 0.41 \\
    \textbf{claude-3-7-sonnet-20250219}            & \textbf{0.83} & \textbf{0.86} & 0.50 & 0.79 \\
    claude-3-5-haiku-20241022                     & 0.74 & 0.83 & 0.07 & 0.60 \\
    gemini-2.5-pro                                 & 0.71 & 0.75 & 0.51 & 0.44 \\
    gemini-2.5-flash                               & 0.71 & 0.71 & 0.57 & 0.68 \\
    gemini-2.5-flash-lite                          & 0.72 & 0.56 & 0.06 & 0.16 \\
    gpt-4.1-2025-04-14                             & 0.79 & 0.78 & \textbf{0.88} & \textbf{0.80} \\
    gpt-4.1-mini-2025-04-14                        & 0.70 & 0.68 & 0.86 & 0.73 \\
    o3-mini-2025-01-31                             & 0.43 & 0.40 & 0.00 & 0.34 \\
    deepseek-r1-distill-llama-70b                  & 0.51 & 0.52 & 0.11 & 0.27 \\
    llama-4-maverick-17b-128e-instruct             & 0.68 & 0.68 & 0.44 & 0.74 \\
    qwen/qwen3-32b                                 & 0.67 & 0.60 & 0.37 & 0.50 \\
    kimi-k2-instruct                               & 0.79 & 0.76 & 0.77 & 0.66 \\
    \bottomrule
  \end{tabular}
  \caption{Stratified F1 scores for each task category (rounded to two decimals).}
  \label{tab:pentestjudge_stratified}
\end{table*}

\subsection{Consistency}
Each grading of the three penetration testing trajectories considered was run five times in order to get an understanding of consistency of judgment. Standard error is computed across five runs per model.

The standard error was less than 1 for all models, representing a maximum $\pm$ 0.065 for classification performance, showing the models are for the most part consistent between runs.

This suggests that using multiple instances of the judge and using the majority vote will not dramatically change the results in the security domain and is unlikely to see increased performance for the price; though this effect may change for very large rubric sizes. 

\section{Conclusion}
We introduce PentestJudge as a comprehensive evaluation methodology for assessing the trajectories of LLM-based security agents. By creating agentic judges that have the ability to closely examine these intermediate outputs, PentestJudge offers the capability to measure qualities necessary for production deployments of such security agents. 

Our results suggest that verification may be less computationally demanding than generation for penetration testing tasks. The trajectories in our study were generated by gpt-4.1, yet models like Kimi-k2-instruct achieved 79\% agreement with human judges at only \$2 per evaluation. Even more striking, budget models like Gemini Flash Lite (F1=0.72) can provide reasonable verification at \$0.17 per trajectory. This contrasts with findings in software engineering domains \cite{tan2025judgebenchbenchmarkevaluatingllmbased} where verification requires models of comparable strength to generation, suggesting the security domain's evaluation requirements may be more tractable. 

The results of failure analyses imply that the results for frontier models would likely be significantly higher with more specifically written requirements, particularly for those that require a cultural or specific technical understanding of security work. This implies that LLM-as-judge can be significantly closer to human performance with the same cost assuming the rubric requirements are more carefully tuned.

We also find that task category is a significant element of grading quality. Operational security tasks, for example, require specific knowledge about how security tools work and what their underlying effect is in the environment that is not inherently obvious in trajectories without underlying domain expertise. This suggests models that are fine-tuned with that information may perform better on these tasks. Adding specific search tools to the PentestJudge harness may be an alternate way of providing more up-to-date information assists in overcoming these model knowledge gaps.

Despite these limitations, these early results represent progress: AI agents succeed in evaluating security agent trajectories across multiple non-trivial steps, suggesting that with harness iteration and model training, these judging systems will continue to improve. This represents a useful tool in the arsenal of security professionals for using their domain expertise to build rubrics that allow them to evaluate deployed security agents at scale beyond what programatically verifiable measures can provide.

\subsection{Future Work}
Future work will focus on improving judges for narrow use-cases. By distilling judges from larger models, it should be possible to reduce their costs in order to make rubric-style evaluation even more attractive for evaluation scenarios.

The output of these judge systems may also act as a reward signal for difficult to verify domains within security, such as respecting operational scope or using particular kinds of tradecraft. These rewards may be used with techniques like GRPO to train models that both achieve operational objectives but also respect operational security and tradecraft requirements.

Finally, while these rubrics can be constructed by human beings in controlled and well-specified environments, the real world is not so well specified. Future work should include determining the feasibility of test-time-generated rubrics in alignment with broad human specifications to encourage the test-time alignment of agents as well as their evaluation in arbitrary security environments.
\newpage

\printbibliography

\vspace{12pt}
\end{document}